\documentclass[conference]{IEEEtran}
\IEEEoverridecommandlockouts
\usepackage{cite}
\usepackage{amsmath,amssymb,amsfonts}
\usepackage{algorithmic}
\usepackage{graphicx}
\usepackage{textcomp}
\usepackage{hyperref}
\def\BibTeX{{\rm B\kern-.05em{\sc i\kern-.025em b}\kern-.08em
    T\kern-.1667em\lower.7ex\hbox{E}\kern-.125emX}}
\begin{document}

\title{Evolutionary Multi-objective Optimization of Real-Time Strategy Micro\\
}

\author{\IEEEauthorblockN{ Rahul Dubey, Joseph Ghantous, Sushil Louis, and Siming Liu}
\IEEEauthorblockA{\textit{Computer Science and Engineering} \\
\textit{University of Nevada}\\
Reno, USA \\
rdubey018@nevada.unr.edu, joseph.ghantous@gmail.com, sushil@cse.unr.edu, simingl@cse.unr.edu}
}

\maketitle

\begin{abstract}
We investigate an evolutionary multi-objective approach to good micro for real-time strategy games. Good micro helps a player win skirmishes and is one of the keys to developing better real-time strategy game play. In prior work, the same multi-objective approach of maximizing damage done while minimizing damage received was used to evolve micro for a group of ranged units versus a group of melee units. We extend this work to consider groups composed from two types of units. Specifically, this paper uses evolutionary multi-objective optimization to generate micro for one group composed from both ranged and melee units versus another group of ranged and melee units. Our micro behavior representation uses influence maps to represent enemy spatial information and potential fields generated from distance, health, and weapons cool down to guide unit movement. Experimental results indicate that our multi-objective approach leads to a Pareto front of diverse high-quality micro encapsulating multiple possible tactics. This range of micro provided by the Pareto front enables a human or AI player to trade-off among short term tactics that better suit the player's longer term strategy - for example, choosing to minimize friendly unit damage at the cost of only lightly damaging the enemy versus maximizing damage to the enemy units at the cost of increased damage to friendly units. We believe that our results indicate the usefulness of potential fields as a representation, and of evolutionary multi-objective optimization as an approach, for generating good micro.

\end{abstract}


\section{Introduction}
Real-Time Strategy games provide difficult challenges for computational intelligence researchers seeking to build artificially intelligent opponents and teammates for such games. In these games, players find and consume resources to build an economy to build an army to defeat an opponent in a series of skirmishes usually culminating in a large decisive battle. Good RTS game play embodies near-optimal sequential decision making in an uncertain environment under resource and time constraints against a deceptive, dynamic, and adaptive opponent (when playing against good players).  Researchers have thus begun focusing on real-time strategy games as a new frontier for computational and artificial intelligence research in games ~\cite{survey}.

RTS game play involves both long-term strategic planning and shorter term tactical and reactive actions. The long-term planning and decision making, often called macromanagement, or just macro for short, can be contrasted with the quick but precise and careful control of game units in order to maximize unit effectiveness on the battlefield.  This short-term control and decision making is often called “micromanagement,” or just micro and good micro can win skirmishes even when a player has fewer units. This paper focuses on evolving good micro for groups of units of different types.

Although much diverse work has been done on generating good micro for RTS games, our work differs in two aspects. First, we use evolutionary multi-objective optimization to tradeoff two objectives: damage done versus damage received. Second, we represent unit behavior using multiple potential fields and an influence map whose parameters evolve to generate micro for groups composed from two types of units. Potential fields of the form ${cx^{e}}$ where $x$ can be distance, health, or weapons cooldown determine unit movement. Influence maps that give high values to map locations with more opponent units specify the location to move towards or to attack. This paper extends earlier work that used the same representation and Evolutionary Multi-Objective Optimization (EMOO) approach in evolving micro for one type of melee unit versus one type of ranged unit ~\cite{ppi}. As in this earlier, we use our own implementation of the NSGA-II algorithms by Deb~\cite{NSGA-II}.

Our results indicate that we can evolve micro for a group of ranged and melee units versus a group of the same number and types of ranged and melee units. The evolved micro performs well against hand selected opponents under a variety of conditions.  Without explicit representation, we see the emergence of kiting behavior for the ranged units, rushing behavior for the melee units, and strong melee units screening for the relatively weak ranged units. The pareto front of evolved solutions contains a variety of tactics suitable for a variety of roles in the broader strategic situation in a particular game. For example, the GA evolves micro that maximizes damage to opponent units while also receiving significant damage, more balanced micro that deals and receives approximately equal amounts of damage, as well as micro that deal little damage but also receives little damage. In the broader picture, this enables a human or AI player to choose the appropriate micro for the current strategic situation. For example, a player may choose micro  that prefers to reduce damage by harassing because it will tend to draw away opponent units from the main force or occupy existing opponent units at a distant location. We believe these results indicate the potential of a multi-objective approach for evolving good micro and to the potential for a potential fields representation of tactical behavior.

The remainder of this paper is organized as follows. Section~\ref{RelatedWork} discusses related work in RTS AI research and common approaches to evolve the micro behavior of units. Section\ref{Simulation Environment} describes our 3D simulation platform, FastEcslent. 
Section\ref{Methodology} introduces the pure potential fields and influence maps that govern micro in simulated skirmishes. This section also describes the NSGA-II algorithm used to evolve the micro behavior. Section\ref{Results and Discussion} presents results and discussion. Finally, the last section draws conclusions from our results and discusses future work.

\section{Related Work} {\label{RelatedWork}}


RTS AI work is popular in both industry and academia. Industry RTS AI developers are more focused on entertainment while academic RTS AI research focuses on learning or reasoning techniques for winning. For example, Ballinger evolved robust build orders in WaterCraft~\cite{ballinger-coevolution}. Gmeiner proposed an evolutionary approach for generating optimal build orders~\cite{opening-strategy1}. Köstler evolve strategies for producing units of one or more types or produce units as quickly as possible~\cite{opening-strategy2}. There is also strong research interest in producing effective group behavior (good micro) in skirmishes since good micro can often turn the tide in close battles. Liu used case-injected genetic algorithm to generate high quality micro~\cite{CIGAR}. Churchill presented a fast search method based on alpha-beta considering duration (ABCD) algorithm for tactical battles in RTS games ~\cite{fas-heuristic}. Again, Liu investigated hill climbers and canonical GAs to evolve micro behaviors in RTS games showing that genetic algorithms were generally better in finding robust, high performance micro~\cite{compare}. Louis and Liu evolved effective micro behavior based on influence maps and potential fields in RTS games~\cite{evolving-effective}. Our paper extends the work in ~\cite{evolving-effective} and represents micro based on influence maps and potential fields for spatial reasoning and unit movement.

In physics, a potential field is usually a distance dependent vector field generated by a force. The concept of artificial potential field was first introduced by Khatib for robot navigation and later this concept was found useful in guiding movement in games~\cite{Khatib}. An influence map structures the world into a 2D or 3D grid and assigns a value to each grid element or cell. Liu compares two different micro representations and the result indicate that even with less domain knowledge the potential fields based representation can evolve a reliable, high quality micro in a three dimensional RTS game~\cite{compare-two-representation}. Schmitt used an evolutionary competitive approach to evolve micro using potential fields based micro representation and results shows that their approach can evolve complex units movement during skirmish~\cite{Schmitt}. 

Early work used influence maps for spatial reasoning to evolve a LagoonCraft RTS game player~\cite{Co-evolution}. 
Sweetser presented an agent which uses cellular automata and  influence maps for decision-making in 3D game environment called EmerGEnt~\cite{Sweetser}. Bergsma proposed a game AI architecture which use influence maps for a turn based strategy game~\cite{Bergsma}. Preuss investigated an evolutionary approach to improve unit movement based on flocking and influence map in the RTS game Glest~\cite{glest}. Uriarte presented an approach to perform kiting behavior using Influence Maps in multi-agent game environment called Nova~\cite{Nova}.

Cooperation and coordination in multi-agent systems, was the focal point of many studies~\cite{19}, ~\cite{20}, ~\cite{21}, ~\cite{22}. Reynolds early work explores an approach to simulate bird flocking by creating a distributed behavioral model that results in artificial agent behavior much like natural flocking ~\cite{Reynolds}.  Similarly Chuang studied controlling large flocks of unmanned vehicles using pairwise potentials ~\cite{Chuang}. 

Within the games community, Yannakakis ~\cite{Yannakakis} evolved opponent behaviors while Doherty~\cite{Doherty} evolved tactical team behavior for teams of agents. Avery used an evolutionary computing algorithm to generate influence map parameters that led to effective group tactics for teams of entities against a fixed opponent~\cite{Avery1, Avery2}. We define potential fields and influence maps in more detail later in the paper. This paper extends Liu \cite{compare-two-representation} and Louis'\cite{ppi} work in dealing with micro for heterogeneous groups of units. 

To run our experiments  we created a simulation model similar to StarCraft called FastEcslent,  our  open  source,  3D,  modular,  RTS  game  environment. The next section introduces this simulation environment in more detail.

\section{Simulation Environment}{\label{Simulation Environment}}

With the release of the StarCraft-II API, StarCraft: Brood War API (BWAPI) and numerous tournament such as Open Real-Time Strategy Game AI Competition, the Artificial Intelligence and Interactive Digital Entertainment StarCraft AI Competition, and the Computational Intelligence and Games StarCraft RTS AI Competition, researchers have been motivated to explore diverse AI approaches in RTS games\cite{SC-champ}. In this work, we ran our experiments in a game simulator called FastEcslent, developed for evolutionary computing research in games\cite{FastEcslent}. Unlike other available RTS-like engines, FastEcslent enables 3D movement, and can run without graphics thus providing simpler integration with evolutionary computing approaches. 

We predefined a set of scenarios where each automated player controls a group of units initially spawned in different locations on  a  map  with  no  obstacles.  The  entities  used  in  FastEcslent reflect   those   in StarCraft,   more   specifically,   Vultures   and Zealots.  A  Vulture  is  a  vulnerable  unit  with  low  hit-points but  high  movement  speed,  a  ranged  weapon,  and  considered effective  when  outmaneuvering  slower  melee  units.  A  Zealot is  a  melee  unit  with  short  attack  range  and  low  movement speed but has high hit-points. Table I shows the details of these properties for both Vultures and Zealots which are used in our experiments. 
\begin{table} {\label{UnitProperties}}
\begin{center}
\caption{Unit Properties Defined In FastEcslent}
\begin{tabular}{|c|c|c|}
\hline
Property & Vulture & Zealot \\
\hline
Hit-points & 80 &160 \\
\hline
MaxSpeed & 64 &40 \\
\hline
MaxDamage & 20 &16*2 \\
\hline
Weapon’s Range & 256 &224 \\
\hline
Weapon’s Cooldown & 1.1 &1.24 \\
\hline
\end{tabular}
\end{center}
\end{table}
Since our research focuses on micro behaviors in skirmishes, we disabled “fog of war” and enabled 3D movement by adding maximum (1000) and minimum (0) altitudes, as well as a climb rate constant, \(r_c\) , of 2. Comparing to StarCraft, units move in 3D by setting a desired heading (\(dh\)), a desired altitude (\(da\)), and a desired speed (\(ds\)). Every time step, a unit tries to achieve the unit’s desired speed by changing its current speed (\(s\)) according to the unit's acceleration ($r_s$).
\begin{equation}
  s = {s \pm r_s\delta t}    
\end{equation}
where \(r_s\) is the unit’s acceleration, $\delta t\ $ is the simulation time step, and $\pm \ $ depends on whether \(ds\) is greater than or less than \(s\). Similarly, 
\begin{equation}
 h = {h \pm r_t\delta t}    
\end{equation}
and
\begin{equation}
a = {a \pm r_c\delta t}  
\end{equation}
Where \(h\) is heading, \(a\) is altitude, \(r_t\) is turn rate, and \(r_c\) is climb rate. From speed, heading, and altitude, we compute 3D unit velocity (vel) and position (pos) as follows: 
\[\textbf{vel} = (s*cos(h),0,s*sin(h))\]
\[\textbf{pos} = \textbf{pos + vel}* \delta t\]
\[\textbf{pos}.y = a\]
Here, bold text indicates vector variables, the $xz$ plane is the horizontal plane, the $y$-coordinate is height, and we assume the unit points along its heading. 

Given a simulation environment within which we can fight battles between unit groups from two different sides, we need an opponent to evolve against. We first describe our representation and then describe how we generate a good opponent to evolve against within this representation. 

%

\section{Methodology}{\label{Methodology}}


We create several game maps (or scenarios) with two types of units on
each side. When we run a fitness evaluation, a decoded chromosome
controls our units as they move, using potential fields, towards a
target location defined by an influence map. This game-simulation
stops when all the units on one side die or time runs out. The
simulation tracks the health of units and provides a multi-objective
fitness (damage done, damage received) for this chromosome to drive
evolution. The rest of this section, describes the scenarios,
potential fields, and influence maps used in our work.

Earlier work has shown that evolving (training) on a single map with
fixed starting locations for all units did not result in robust
micro\cite{evolving-effective}. We therefore train our units over five
different scenarios and measure the robustness of evolved micro on 50
unseen randomly generated scenarios. In this work, randomly generated
scenarios means only that units start at different initial positions
at the beginning of a fitness evaluation. Scenarios are constructed
from "clumps" and "clouds" of entities; defined by a center and a
radius. All units in a clump are distributed randomly within a sphere
defined by this radius ($400$ for this paper).  Units in a cloud are
distributed randomly within $10$ units of the sphere boundary defined
by the center and radius (also $400$).

We created two sides; player1 with $5$ Vultures and $5$ Zealots and
player2 with $5$ Vultures and $5$ Zealots. The training scenarios are
as follows: (a) A clump of player1 versus a clump of player2, (b) A
clump of player1 units surrounded by a cloud of player2 units, (c) A
clump of player2 units surrounded by a cloud of player1 units, (d) A
set of player1 units within range of $250$ in all three dimension
centered at the origin and a set of player2 units within $250$ in all
three dimension centered at $650$, and (e) the same distributions of
units but with the players swapping their centers. Our evaluation
function ran each of these five scenarios for every chromosome during
fitness evaluation and the value returned by the simulation for each
objective is averaged over these scenarios. This results in evolving
more reliable micro that can do well under different training
scenarios.

Once a scenario starts running, units have to come up with a target
location to attack. An influence map determines this target location.


\subsection{Influence Maps}
A typical IM is a grid defining spatial information in a game world, with values assigned to each grid-cell by an IMFunction. These grid-cell value are computed by summing the influence of all units within range, $r$ of the cell. $r$ is measured in number of cells. The IM not only considers units’ positions in the game world but also includes the hit-points and weapon cool-down of each unit. The influence of a unit at the cell occupied by the unit is computed as the weighted linear sum these factors. A unit's influence thus starts as this weighted linear sum at the unit's cell and decreases with distance from this cell by a factor: $I_f$. The NSGA evolves these parameters and evolving units move towards the lowest IM grid-cell value~\cite{ppi} using potential fields to guide all movement.

\subsection{Potential Fields} 

We use potential fields to guide unit movement to the target location
provided by the IM. Once near the opponent, we would like our units to
maneuver well based on the location of enemy units, their health, and
the state of their weapons. We thus define and use attractive and
repulsive potential fields for each of these factors. Since the fields
for friendly units should be different from the fields for enemy
units, we use two such sets of potential fields. Finally, the target
location also exerts an attractive potential. This results in a total
of $2$ (attractive, repulsive) $\times 3$ (location, health, weapons
state) $\times 2$ (friend, enemy) $+ 1$ (target) $ = 13$ potential
fields for guiding one type of unit's movement against an enemy also
composed of only one type of unit. We use the same techniques
from~\cite{ppi} to convert the vector sum of these potential fields
into a desired heading and desired speed and same ranges of value for
potential field parameters.

\begin{figure}
\centerline{\includegraphics[scale=0.55]{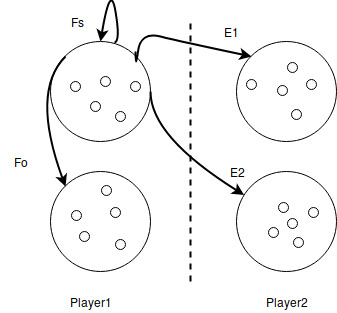}}
\caption{Potential fields needed for groups composed from two types of units.}
\label{PPFForTwoTypesOfUnits}
\end{figure}

Once we move to micro for groups composed from two types of units, the
number of potentials fields increases. Figure~\ref{PPFForTwoTypesOfUnits} shows the four
sets of potential fields needed when dealing with groups composed from
two types of units. Instead of one set of potential fields for friends
and one set of potential fields for enemies, we will need two sets of
potential fields corresponding to the two types of friendly units, and
two sets of potentials fields for the two types of enemy units. Each
type of friendly unit can then respond differently to the two types of
friendly units and differently to both types of enemy units. Results
show that these potential fields enable the evolution of high
performance micro.
%



We can see that Equation~\ref{growthEquation} gives the total number
of parameters required to deal with $n$ different types of units to a
side. A total $12$ (\(p\)) attraction and repulsion potential fields
are required for one type of friend and enemy units with $2$
parameters per potential field. As different types of units are added
each side, few parameters are counted multiple time such as potential
field generated by distance between two different types of units each
side. Summation in Equation~\ref{growthEquation} subtracts extra
counted parameters. \(q\) is a constant represents target attraction
potential and IM parameters. Thus for two different types of units
each side, total 106 number of parameters required.
\begin{equation}{\label{growthEquation}}
\mbox{Number of parameters} = (q + 2*p*n)*n - \displaystyle{ \sum_{i \in n}4*(i-1)}
\end{equation}



These parameters provide a target location and guide unit movement
towards the target. If enemy units come within weapons range of a
friendly unit, the friendly unit targets the nearest enemy unit. In
our game simulation all entities can fire in any direction even while
moving from one location to other. With a good set of parameters the
units evolve effective micro that tries to maximize damage done to
enemy units while minimizing damage taken.

Although some work has combined damage done and damage received into
one objective to be maximized, we keep the objectives separated and
use an evolutionary multi-objective optimization approach to evolve a
diverse pareto front. Specifically, we use our implementation of the
Fast Non-dominated Sorting Genetic Algorithm (NSGA-II) to evolve a
pareto front of micro behaviors for heterogenous groups composed from
two types of units. We try to maximize damage done to enemy units
while minimizing damage to friendly units. Assume that we normalize
damage done and damage received to span the range $[0..1]$,
Equation~\ref{multi-objective} describes our multi-objective optimization problem.
\begin{equation}{
    \label{multi-objective}}
  \mbox{Maximize} \bigg[  \sum_{enemies}(D_e) ,  \sum_{friends}(1-D_f) \bigg]
\end{equation}
Here, \(D_e\) represents damage done to enemy units and \(D_f\)
represents damage to friendly units. To minimize damage to friendly
units we subtract from the maximum damage possible, $1$, to also turn
the second objective into a maximization objective.  This normalized,
two-objective fitness function used within our NSGA-II implementation
then produces the results described in our results section. 



\subsection{Baseline Opponent AI} 

In order to produce high quality micro behavior, finding a good
opponent to play against is crucial. Instead of handcoding an
opponent, we use a two step approach to find a good opponent. First,
we generated 30 random chromosomes that we used as opponents and ran
NSGA-II against each one of them with population size of 20 for 30
generations. The best opponent is the one that does most damage to
friendly units. We thus choose the opponent chromosome that does the
most damage as the next opponent. We then run our NSGA against this
chromosome and choose the most balanced, closest to (0.5, 0.5),
resulting chromosome from the last generation pareto front as the next
opponent. We repeat this process five times (five steps).

\begin{figure}[htbp]
  \centerline{
    \includegraphics[width=3.5in]{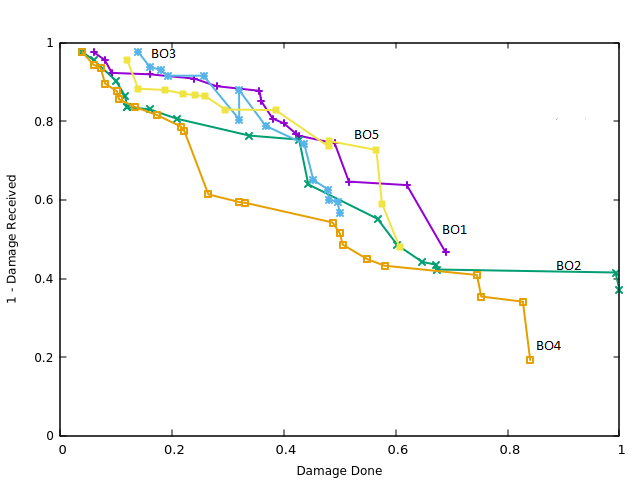}
  }
\caption{Pareto front of 1000 random chromosome against BO1 to BO5}
\label{Opponent}
\end{figure}

\begin{figure}[htbp]
  \centerline{
    \includegraphics[width=3.5in]{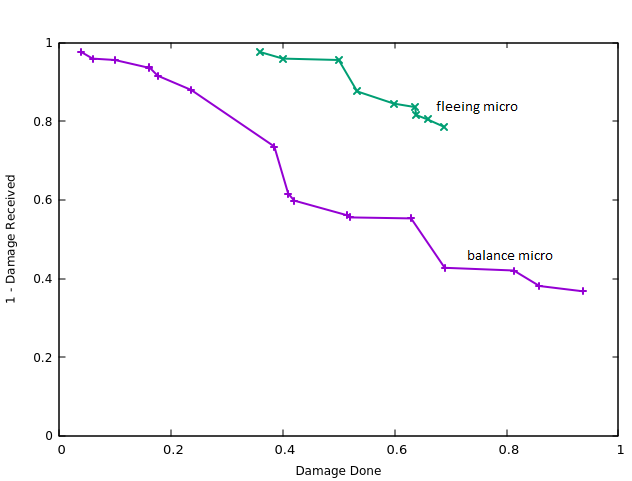}
  }
\caption{Comparing the pareto front of 3750 random chromosome against good balanced and good fleeing micro}
\label{BO4Flee}
\end{figure}

Figure~\ref{Opponent} shows the performance of $1000$ randomly
generated chromosomes against the balanced individual from the last
generation for each of the five steps above.
The line maked BO$i$ represents the pareto front of these $1000$
random chromosomes against the best balanced individual in the
$i^{th}$ step. The x-axis represents damage done, while the y-axis
represents 1 - damage received. The point $(1, 1)$ then represents
micro that destroys all enemy units and receives no damage. $(1, 0)$
is micro that does destroys all opponents but also loses all friendly
units. $(0, 1)$ usually indicates fleeing behavior, friendly units
deal no damage and receive no damage. $(0, 0)$ is bad, friendly units
did no damage and received maximal damage - micro to be avoided. From
the figure, we can see that the $1000$ chromosomes did worst against
BO4. Finally, to confirm that BO4 would make a good opponent to evolve
against, we picked $3750$ random chromosomes and compared BO4 against
antother individual from the step four pareto front. $3750$ comes from
our population of $50$ multiplied by the $75$ generations we
run. Figure~\ref{BO4Flee} shows how these two individuals fare against
these new random chromosomes. Clearly these individuals perform worse
against BO4 and we thus chose BO4 as our opponent in the experiments
described below.

\section{Results and Discussion}{\label{Results and Discussion}}

We use real-coded parameter with simulated binary crossover (SBX)
along with polynomial mutation. After experimenting with different
values, we settled on the following. Crossover and mutation
distribution indexes were both set to $20$. We used high probabilities
of crossover ($0.9$) and mutation ($0.05$) to drive diversity.


\subsection{Pareto Front Evolution of Final Experiment}
\begin{figure}[htbp]
\centerline{\includegraphics[scale=0.55]{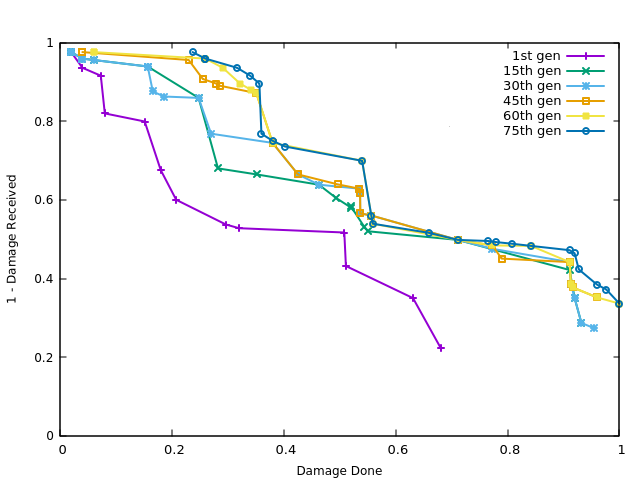}}
\caption{Micro Evolution for Friendly Unit in Final Experiment}
\label{FriendlyUnits}
\end{figure}

We evolved micro for groups of $5$ vultures and $5$ zealots versus an
identical opponent also with $5$ vultures and $5$ zealots.  We used a
population size of $50$ for $75$ generations and report results over
$10$ runs using a different random seed for each run. Figure~\ref{FriendlyUnits} shows
the evolution of the pareto front at intervals of fifteen ($15$)
generations for one run of our parallelized-evaluation
NSGA-II. Broadly speaking, the pareto front moves towards $(1, 1)$
while maintaining representatives along the tradeoff curve for
maximizing damage done and minimizing damage received. We can see the
maintainence of a diverse set of micro making a diverse set of
tradeoffs between damage done and received. These results provide
evidence that we can evolve a diverse set of micro tactics that learns
to performs well against an existing opponent

\begin{figure}[htbp]
  \centerline{
    \includegraphics[width=3.5in]{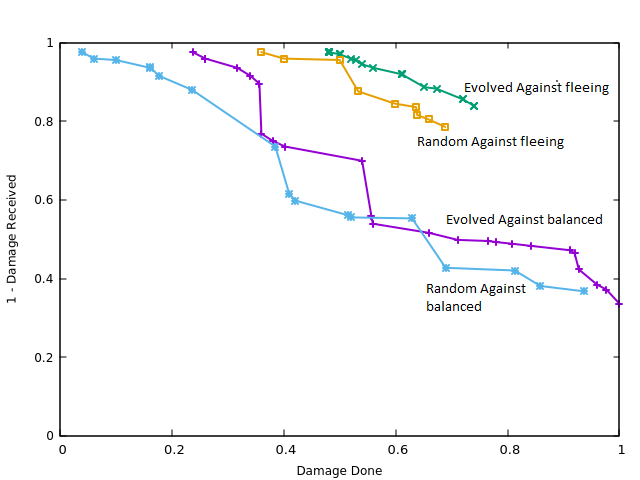}
  }
\caption{Comparing evolved micro against 3750 random chromosomes}
\label{BO4VSEvolved}
\end{figure}

To test the effectiveness of our evolutionary multi-objective optimization approach, we played a balanced
individual and a fleeing individual from the $75^{th}$ generation pareto front against $3750$
randomly generated chromosomes. Figure~\ref{BO4VSEvolved} shows how
these random chromosomes did against the evolved micro. For
comparision, we also ran BO4 and the fleeing micro from our opponent
evolution experiments against these random individuals.
The figure shows that our evolved balanced individual does better than
BO4, and the evolved fleer also does better than the step four
fleer. Evolutionary multi-objective optimization's attention to
producing a diverse set of solutions along the pareto front leads seems to
lead to robust micro. Watching the micro we can see the emergence of
kiting, withdrawing, and other kinds of behavior often seen in human
game play. Videos of the evolved micro are available at
\url{https://www.cse.unr.edu/~rahuld/Experiment/}.

\begin{figure}[htbp]
  \centerline{
    \includegraphics[scale=0.55]{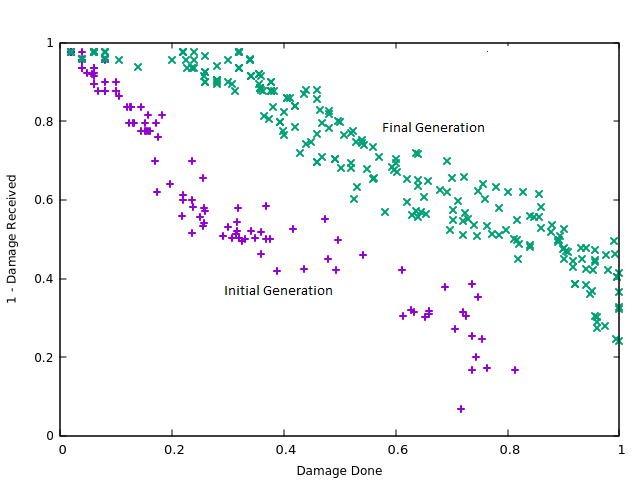}
  }
\caption{The initial and final generation pareto front over ten runs for evolved micro}
\label{ParetoFrontFromAllTenRuns}
\end{figure}

Figure~\ref{ParetoFrontFromAllTenRuns} plots the combined pareto front
in the first generation over all ten random seeds versus the combined
pareto front in the last generation over the ten random seeds. That
is, we first did a set union of the pareto fronts in the ten initial
randomly generated populations. The points in this union over all ten
runs are displayed
as purple $+$ for the initial generation (generation $1$) points and
as green $\times$ for the points in the final generation (generation
$75$). The figure then shows progress between the first and last
generation over all ten runs. 
We can see that the last generation pareto front produces micro on
one extreme on the left (0.02, 0.98) representing a strong tendency to flee, to the other extreme on the right (1,0.25) denoting an
aggressive attacking micro behavior. There are a number of solutions near the middle with balanced micro behavior.

\begin{figure}[htbp]
  \centerline{
    \includegraphics[width=3.5in]{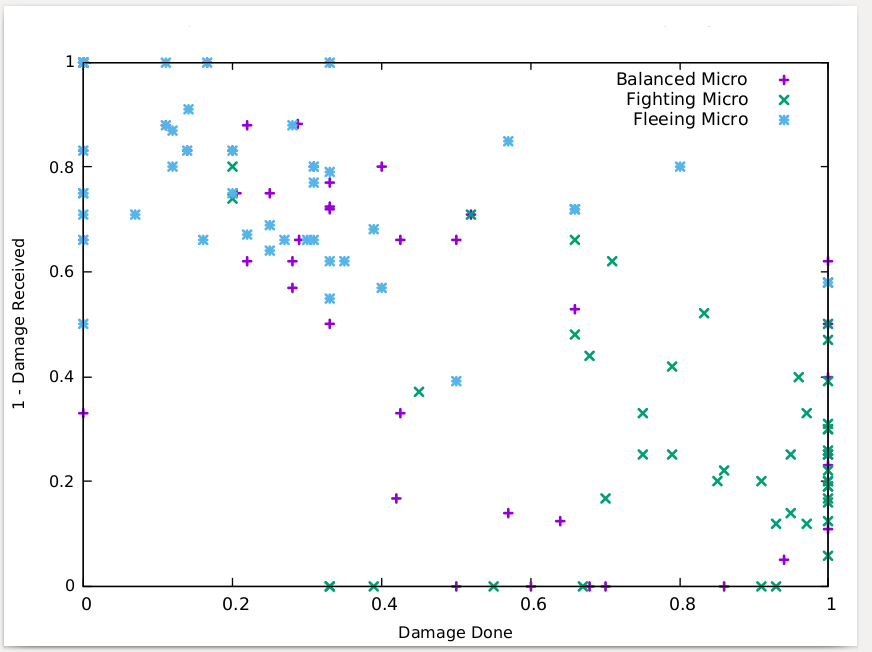}
  }
\caption{Robustness of evolved micro on 50 random testing scenarios}
\label{Robustness}
\end{figure}

To further check the robustness of our evolved micro on the last
generation pareto front, we decided to select one balanced, one
fleeing, and on attacking example of micro from this last generation
and play against BO4 in a variety of different randomly generated
scenarios. In these $50$ scenarios, we randomly varied the numbers of
zealots and vultures, both between $5 - 10$, and made sure that both
sides had identical units.  Figure~\ref{Robustness} shows results,
indicating that
the evolved attacking micro (green $\times$s) comes in on the lower right, generally dealing damage while also receiving significant damage.
On average over the $50$ scenarios, the attacking micro leads to objective function values of $(0.812, 0.291)$, while the balanced micro leads to an average of $(0.39, 0.59)$ and the fleeing micro's average fitness comes to $(0.21, 0.79)$. 



Finally, we played the evolved attacking micro against larger numbers of
opponents. The video at
\href{https://www.cse.unr.edu/~rahuld/Experiment/video1}{https://www.cse.unr.edu/$\sim$rahuld/Experiment/video1}
shows how $5$ vultures and $5$ zealots controlled by our evolved
attacking micro plays against and defeats $5$ vultures and $10$
zealots controlled by BO4. The attacking micro manages to destroy all
$15$ opponent units showing that better micro can win even when outnumbered. A
second video at
\href{https://www.cse.unr.edu/~rahuld/Experiment/video2}{https://www.cse.unr.edu/$\sim$rahuld/Experiment/video2}
shows our attack micro controlled $5$ vultures and $5$ zealots playing
against $5$ Vultures and $15$ Zealots controlled by BO4. This is an example of the type of effective kiting that evolves over time.


\section{Conclusion}

This paper focused on extending research in multi-objective
optimization and potential fields based representation to evolve micro
for groups composed from heterogeneous (two) types of units.  We
choose a group of ranged and melee unit to play against a group of
ranged and melee considering damage done and damage received as two
objective functions. We use an evolutionary multi-objective
optimization approach that maximizes damage done and minimizes damage
received to tune influence map and potential field parameter values
that lead to winning skirmishes in our scenario.

We can see the emergence of kiting and other complex behavior as the
poulation evolves. The multi-objective problem formulation the fast
non-dominated sorting GA evolve pareto fronts that produced a diverse
range of micro behaviors. These solutions not only beat the opponent
that they played against to determine fitness, but are robust to
different numbers of opponents and can beat an opponent even when outnumbered. 

Although this work dealt with two unit types, we would like to extend
our work to multiple unit types and to reduce the need for a good
opponent to evolve against. Since we had to manually co-evolve the
opponent in this paper, we plan to investigate coevolutionary
multi-objective approaches. We would like to use a multi-objective,
co-evolutionary algorithm to co-evolve a range of micro that is robust
against a range of opposition micro. Second; we plan to work on the
StarCrat -II API to implement our approach and representation to
evolve good micro to test against human experts.

\bibliographystyle{unsrt}
\bibliography{main}

\end{document}